\documentclass{article}
\usepackage{spconf,amsmath,graphicx}
\usepackage{amssymb,bm}
\usepackage{subfigure}
\usepackage{caption}
\usepackage{tabularx}
\graphicspath{{images/}}

\title{Sparse Spatial Attention Network for Semantic Segmentation}
\name{Mengyu Liu and Hujun Yin}
\address{Department of Electrical and Electronic Engineering, The University of Manchester, Manchester, UK}
\begin{document}
%
\maketitle
\begin{abstract}
The spatial attention mechanism captures long-range dependencies by aggregating global contextual information to each query location, which is beneficial for semantic segmentation. In this paper, we present a sparse spatial attention network (SSANet) to improve the efficiency of the spatial attention mechanism without sacrificing the performance. Specifically, a sparse non-local (SNL) block is proposed to sample a subset of $key$ and $value$ elements for each query element to capture long-range relations adaptively and generate a sparse affinity matrix to aggregate contextual information efficiently. Experimental results show that the proposed approach outperforms other context aggregation methods and achieves state-of-the-art performance on the Cityscapes, PASCAL Context and ADE20K datasets.   
\end{abstract}
\section{Introduction}
\label{sec:intro}

Semantic segmentation, a process of labelling all pixels of an image to various classes based on their properties, is a fundamental topic in computer vision. In recent years, deep networks, esp. convolutional neural networks (CNNs) have elevated the performance of semantic segmentation algorithms to new heights \cite{long2015fully, ronneberger2015, chen2018encoder, liu2015parsenet, zhao2017}. It has been shown that encoding rich contextual information helps achieve good results in semantic segmentation \cite{zhao2017, chen2018deeplab, liu2015parsenet, fu2019dual}. However, due to the local connectivity of CNN layers, the size of receptive field is often too small to exploit sufficient features. Even in recent deep models, effective receptive field is still not large enough to cover the entire image \cite{zhou2014object}. 

To address this problem, many methods have been proposed to aggregate global contextual information from local features. Some \cite{chen2018deeplab, zhao2017, peng2017, yang2018} utilize large filters, atrous spatial pyramid pooling (ASPP) or pyramid pooling module (PPM) to enlarge receptive fields and extract multi-scale features at the end of the network. Others \cite{wang2018non, fu2019dual, zhao2018psanet} adopt spatial attention mechanism to capture long-range dependencies within the feature maps. The non-local block proposed in \cite{wang2018non} has been widely applied for semantic segmentation as an instance of spatial attention mechanism to aggregate global contextual information densely. For each query element in a non-local block, all key elements are used to compute the pairwise relations with the query element and generate a dense affinity matrix. Then the contextual information at all locations is aggregated by a weighted sum with the weights defined by the affinity matrix. Although the non-local block improves the performance significantly, it requires high computational costs on high-end GPU-based platforms \cite{zhu2019asymmetric, huang2019ccnet}. Moreover, it is worth noting that some features may contain irrelevant information, thus aggregating these features is useless or even harmful to the performance.   

Based on these observations, we propose a sparse non-local (SNL) block to aggregate contextual information from a sparse set of features, reducing the computational cost without sacrificing the performance. For each query element in the proposed SNL block, only a few key and value elements are sampled to compute a small sparse affinity matrix, and the sampling locations are learnable. To demonstrate the effectiveness of the proposed SNL block, we build a sparse spatial attention network (SSANet) based on the ResNet-FCN backbone with the SNL block integrated. The proposed network has been evaluated on various semantic segmentation benchmarks, including the Cityscapes \cite{cordts2016}, the PASCAL Context \cite{mottaghi2014role} and the ADE20K \cite{zhou2019semantic}, and achieved state-of-the-art performances.

\section{Related Work}

\noindent\textbf{Semantic Segmentation.} Long et al. proposed the fully convolutional network \cite{long2015fully} where the fully connected layers are replaced with convolutional layers to convert the semantic segmentation task into a pixel-level classification task. Chen et al. proposed a group of segmentation networks called DeepLab \cite{chen2018deeplab, chen2017rethinking, chen2018encoder}, which adopted atrous (dilated) convolutions to increase receptive fields and preserve spatial resolution of feature maps simultaneously. Zhao et al. \cite{zhao2017} utilized a group of average pooling operations to encode multi-scale features. Another popular method is the encoder-decoder structure \cite{badrinarayanan2015, ronneberger2015}, composed of an encoder and a symmetric decoder, where different level features are extracted in each stage of the encoder and the image resolution is recovered in the decoder step by step. 

\noindent\textbf{Contextual Information.} Exploring contextual information helps improve performance in semantic segmentation due to multiple scales of objects in images and long-range dependencies between different locations. In \cite{chen2018deeplab}, an atrous spatial pyramid pooling (ASPP) module was proposed to encode multi-scale contextual features by several parallel atrous convolutions of different rates. While in \cite{zhao2017}, a pyramid pooling module was adopted at the end of the model to exploit multi-level contextual information by pooling operations.

\noindent\textbf{Spatial Attention.} Spatial attention mechanism can be considered as a contextual feature aggregation method, which models the long-range dependencies between each pair of elements in the feature maps and then gathers contextual information to each location. Wang et al. \cite{wang2018non} proposed the non-local block and integrated it in deep networks as a self-attention module to capture long-range dependencies for video classification. In \cite{fu2019dual}, a spatial and a channel attention modules were adopted to capture relations and re-weight each location and channel, respectively. Zhao et al. \cite{zhao2018psanet} proposed a two-branch attention block to collect and distribute information separately.    

\section{Methods}

\subsection{Non-local Block and Spatial Attention}  
\label{sec:non-local}
\begin{figure}[htb]
	\centering    	
	\subfigure[Non-local block]{
		\includegraphics[width=0.98\linewidth]{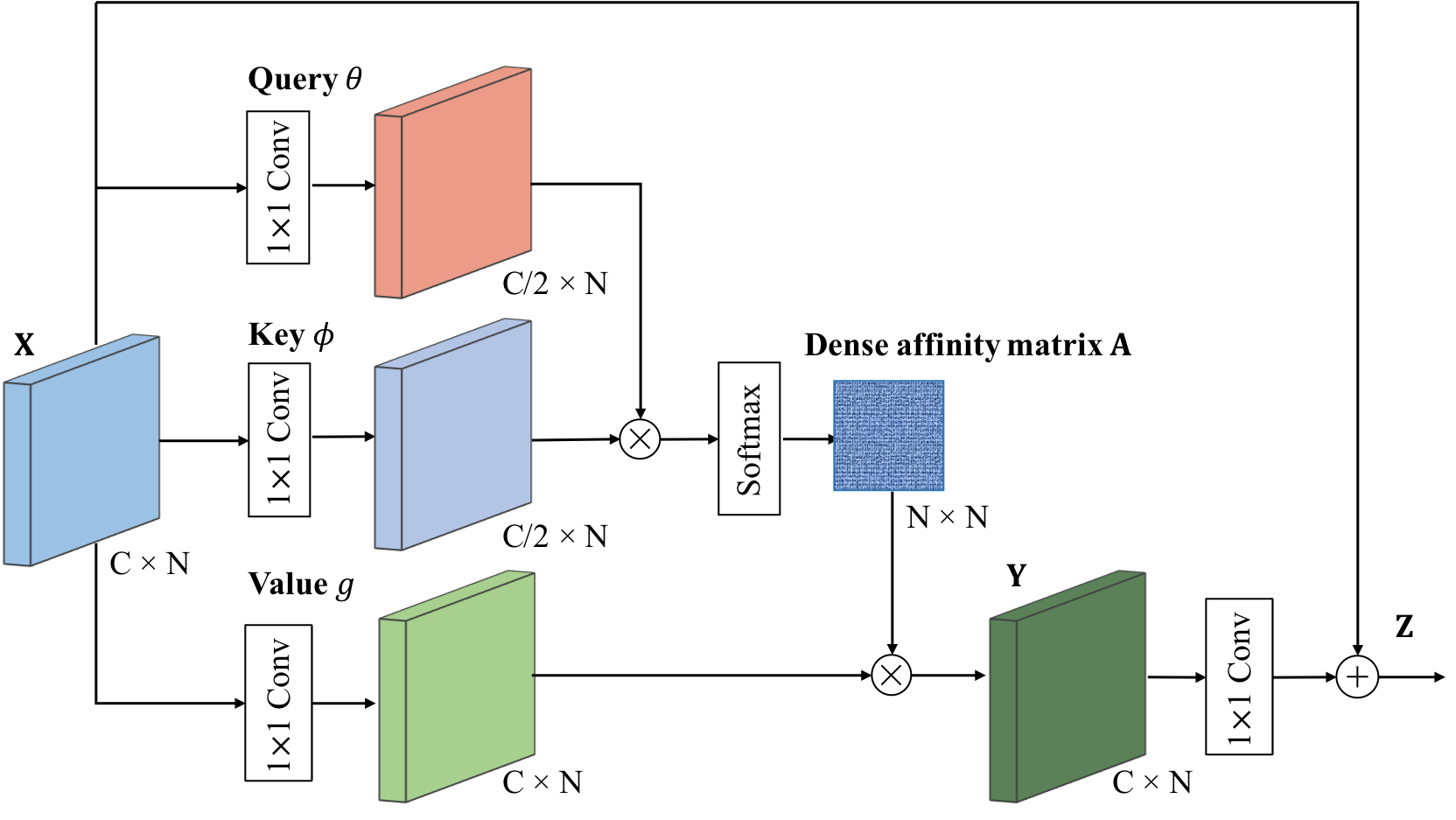}
		\label{fig:nonlocal}}	
	\subfigure[Sparse non-local block]{
		\includegraphics[width=0.98\linewidth]{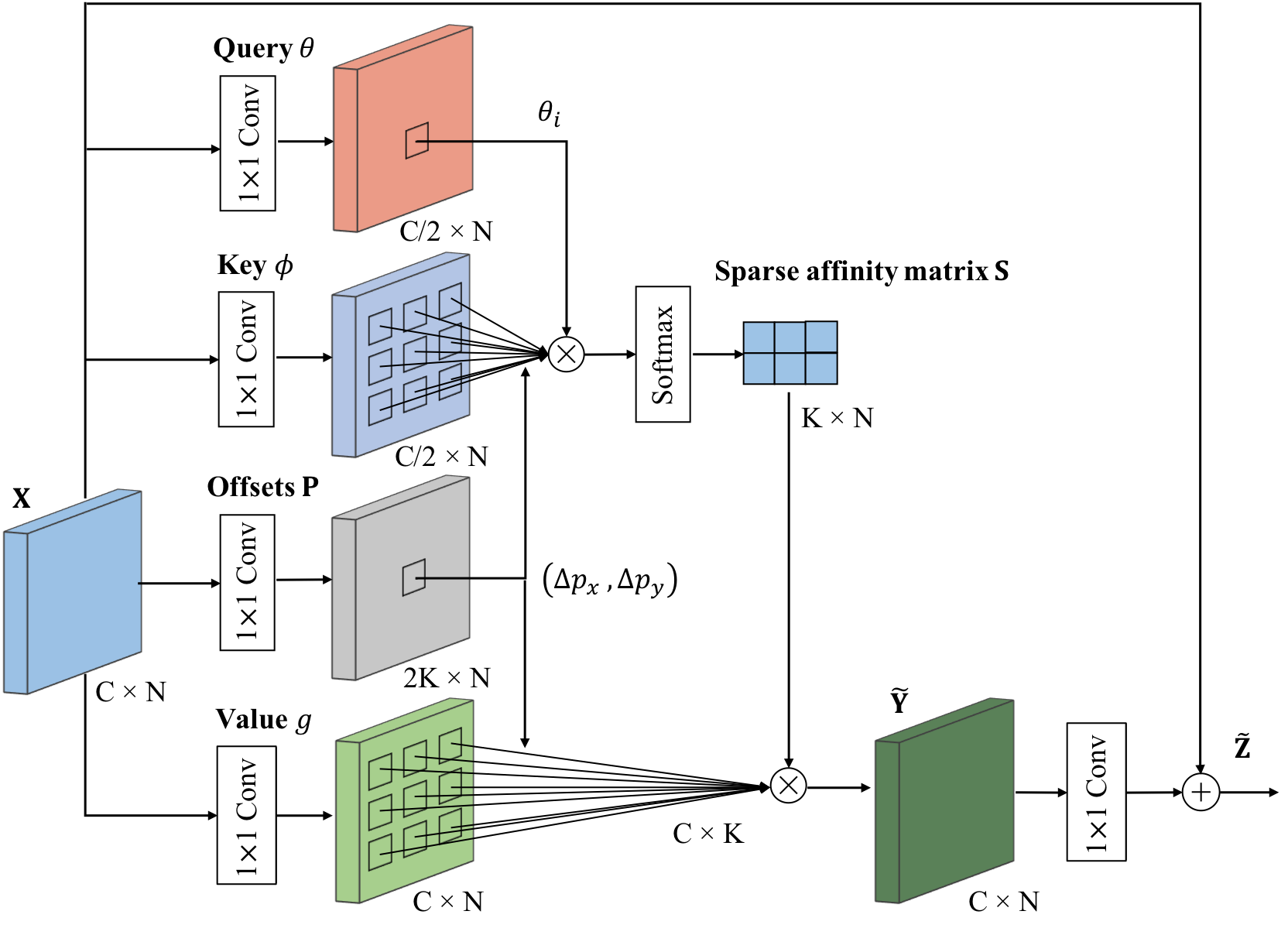}
		\label{fig:snl}}	
	\caption{Structures of (a) non-local block and (b) proposed sparse non-local block, where $\otimes$ is matrix multiplication, $\oplus$ is element-wise addition, $C$ is the channel number, $N$ is the spatial size of the feature map, $K$ is the number of sampled key and value elements for each query element, and $K \ll N$.}
	\label{fig:nonsnl}
\end{figure}

The structure of the non-local block \cite{wang2018non} is depicted in Fig.~\ref{fig:nonlocal}. For a given input feature map $\boldsymbol{X} = [\boldsymbol{x}_1, \cdots, \boldsymbol{x}_N] \in \mathbb{R}^{C \times N}$, where $C$ and $N$  are the channel number and spatial size of the feature map, respectively. Three $1\times1$ convolutions: $\theta$, $\phi$ and $g$ are used to transform $\boldsymbol{X}$ to the query, key and value embeddings respectively, i.e.  $\theta(\boldsymbol{X}), \phi(\boldsymbol{X}) \in \mathbb{R}^{C/2 \times N}$ and $g(\boldsymbol{X}) \in \mathbb{R}^{C \times N}$. Then a dense affinity matrix, $\boldsymbol{A} \in \mathbb{R}^{N \times N}$, is computed by a matrix multiplication and softmax normalization as
\begin{equation}
a_{ij} = \frac{{\rm exp}(\theta(\boldsymbol{x}_i)^{\top} \phi(\boldsymbol{x}_j))}{\sum_{t=1}^{N}{{\rm exp}(\theta(\boldsymbol{x}_i)^{\top} \phi(\boldsymbol{x}_t))}},
\label{matrix}
\end{equation}           
where $a_{ij} \in \boldsymbol{A}$ is the normalized similarity between $\theta(\boldsymbol{x}_i)$ and $\phi(\boldsymbol{x}_j)$, and $i$, $j$ and $t$ are spatial location indexes of query and key respectively. After generating the affinity matrix, another matrix multiplication is used to aggregate the contextual information with value contents as follows,
\begin{equation}
\boldsymbol{Y} = g(\boldsymbol{X}) \boldsymbol{A}^{\top}. 
\label{attention}
\end{equation}
The final output $\boldsymbol{Z}$ can be computed as 
\begin{equation}
\boldsymbol{Z} = \gamma(\boldsymbol{Y}) + \boldsymbol{X}, 
\label{output}
\end{equation}
where $\gamma$ refers to a $1\times1$ convolution. In such scenario, long-range dependencies are captured and used to provide rich contextual information. 

\subsection{Sparse Non-local Block}  
Although the non-local block has proven effective for semantic segmentation \cite{fu2019dual, zhu2019empirical}, it is very time consuming. The review of non-local block in Sec.~\ref{sec:non-local} indicates that the matrix multiplications in Eqn.~\ref{matrix} and Eqn.~\ref{attention} dominate most of the computational resources, and the computational complexity is $\mathcal{O}(N^2C)$. With this consideration, if the calculation of the dense affinity matrix could be replaced with a small sparse affinity matrix, the computational complexity would be reduced.      

In the proposed sparse non-local (SNL) block, we sample a subset of key elements to multiply with each query element and generate a sparse affinity matrix denoted as $\boldsymbol{S}\in \mathbb{R}^{N \times K}$, where $K$ is the number of sampled key elements. The sampled key elements should be representative and the sampling region should cover large area to encode contextual information, and the number of sampled elements should also be small to reduce the computational complexity. 

Motivated by the deformable convolution \cite{dai2017deformable}, we restrict the initial sampling locations to the neighbouring locations of the corresponding query pixel, and then shift the locations with offsets $\boldsymbol{P} \in \mathbb{R}^{2K \times N}$, which is learned by a convolution performed on $\boldsymbol{X}$, and the channel dimension $2K$ corresponds to $K$ 2D offsets. Hence, the final sampling coordinates $(t_x, t_y)$ of each key element are calculated as
\begin{equation}
t_x = p_x + \Delta p_x, \qquad t_y = p_y + \Delta p_y,
\label{coordinates}
\end{equation} 
where $(p_x, p_y)$ and  $(\Delta p_x, \Delta p_y)$ denote the the initial coordinates and the corresponding offsets in $\boldsymbol{P}$, respectively. As the offsets are typically fractional, bilinear interpolation is used to sample key values and make the offsets computation differentiable. The sampled key element $\overline{\phi(\boldsymbol{X})}$ is calculated as
\begin{equation}
\overline{\phi(\boldsymbol{X})} = G(\phi(\boldsymbol{X}), (t_x, t_y)) ,
\label{sampled}
\end{equation} 
where $G$ is the bilinear interpolation function. For $\theta(\boldsymbol{x}_i)$ and $\overline{\phi(\boldsymbol{x}_j)}$, the pairwise similarity $s_{ij} \in \boldsymbol{S} $ can be computed as
\begin{equation}
s_{ij} = \frac{{\rm exp}(\theta(\boldsymbol{x}_i)^{\top} \overline{\phi(\boldsymbol{x}_j)})}{\sum_{t=1}^K {{\rm exp}(\theta(\boldsymbol{x}_i)^{\top}\overline{\phi(\boldsymbol{x}_t)})}}.
\label{sparse matrix}
\end{equation} 
Afterwards, the value contents $\overline{g(\boldsymbol{X})} \in \mathbb{R}^{C \times K}$ are sampled at the same locations as $\overline{\phi(\boldsymbol{X})}$, and the contextual information is aggregated as
\begin{equation}
\begin{aligned} 
\widetilde{\boldsymbol{Y}} &= \overline{g(\boldsymbol{X})} \boldsymbol{S}^{\top}, \\
\widetilde{\boldsymbol{Z}} &= \gamma(\widetilde{\boldsymbol{Y}}) + \boldsymbol{X}. 
\label{sparse attention}
\end{aligned} 
\end{equation}  

Because of replacing the dense matrix multiplications in the non-local block with sparse matrix multiplications defined by Eqn.~\ref{sparse matrix} and Eqn.~\ref{sparse attention}, the computational complexity is reduced to $\mathcal{O}(NKC)$, which is significantly lower than $\mathcal{O}(N^2C)$ with $K \ll N$\footnote{$N$ is 2401 for an input feature map of spatial size $49\times49$ when evaluating on cityscapes dataset, and $K$ is 81 in our model.}. 

The structure of SNL block is depicted in Fig.~\ref{fig:snl}. First, four $1\times1$ convolutions are performed on the input features to generate query, key, value and offsets, respectively. Next, $K$ key elements are sampled for each query element based on the offsets, and then are multiplied by the corresponding query element and normalized to generate the sparse affinity matrix. Then, the generated matrix is multiplied with the value features to obtain $\widetilde{\boldsymbol{Y}}$. Finally, a $1\times1$ convolution is applied for feature fusion. 

\begin{figure}[t!]
	\centering    
	\includegraphics[width=0.98\linewidth]{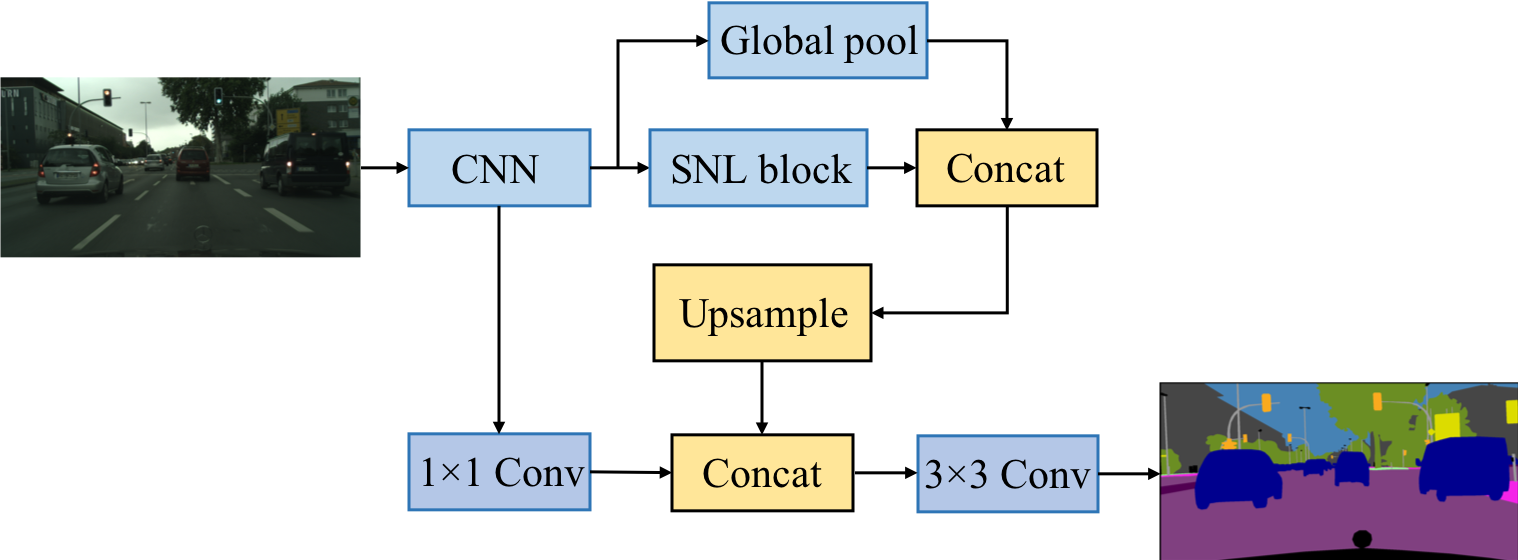}		
	\caption{Architecture of proposed SSANet.}
	\label{fig:Network}
\end{figure}

\subsection{Network Architecture}
The entire network architecture is shown in Fig.~\ref{fig:Network}. We use MobileNetV2 \cite{sandler2018} and ResNet-101 \cite{he2016} as our backbones. Following the previous studies \cite{chen2018encoder}, we remove the last downsampling operation in the backbones and utilize dilated convolutions in the last stage to maintain the size of receptive field. We reduce the number of channels to 512 in ResNet-101-based network and 256 in MobileNetV2-based network with a $3\times3$ convolution, and then we apply the SNL block on the reduced features. Besides, a global average pooling branch is introduced
and concatenated with the reduced features to provide image-level information and improve the performance. Finally, we also employ a simple decoder. Output features of the encoder part are first upsampled, and then concatenated with the reduced low-level features from stage 2 in the backbone. After the concatenation, two $3\times3$ convolutions are used to fuse features before the final $1\times1$ classifier.        

\section{Experiments}

\subsection{Implementations and Datasets}
We conducted all the experiments based on PyTorch \cite{paszke2017automatic} with CUDA 10.0. We employed ImageNet-pretrained \cite{deng2009imagenet} MobileNetV2 and ResNet-101 as the backbones, and stochastic gradient descent (SGD) algorithm with momentum 0.9 and weight decay 0.0001 was used to train the networks on all the datasets. The ``poly'' learning rate policy \cite{liu2015parsenet} (the learning rate is multiplied by $(1-\frac{iter}{max\_iter})^{power}$ with $power=0.9$) was used, and the initial learning rate was set to 0.005. We employed random horizontal flipping, random scaling and random cropping for data augmentation. The mean Intersection-over-Union (IoU) metric was used as the evaluation metric. 

\noindent\textbf{Cityscapes.} The dataset contains 5000 finely annotated images. It is divided into 2975, 500 and 1525 for training, validation and test, respectively. All images are of size $2048 \times 1024$, and 19 semantic classes are used for segmentation.

\noindent\textbf{PASCAL Context.} The dataset provides dense semantic annotations for the PASCAL VOC 2010 images and contains 4998 training and 5105 validation images. Following the previous work \cite{yuan2019object}, we used the 59 classes annotations to train the networks and to measure the performances. 

\noindent\textbf{ADE20K.} The dataset is a challenging segmentation dataset which contains 150 semantic classes. The training and validation sets consist of 20K and 2K images respectively.

\subsection{Ablation Studies}   
To evaluate the performance of the proposed method, we first conducted ablation studies on the Cityscapes validation set with single scale testing.

\noindent\textbf{Ablation on sampling number.} We adopted a simple MobileNetV2-based FCN as the baseline, and it achieved 70.39\% mean IoU result. Then we integrated the sparse non-local block at the end of backbone. We adopted different schemes to evaluate the effect of the SNL block by changing the number of sampled key and value elements for each query element. Results are shown in Table~\ref{table:SNL study}, indicating that the sparse non-local block gave better results. Specifically, the improvements comparing with the baseline increased with sampling more key elements, indicating that encoding more contextual information are beneficial for segmentation task. When sampling number was larger than 81, the improvements began to drop, this may due to that some key elements contain redundant or unhelpful information, leading to distortion. Hence we selected 81 as the sampling number in the final architecture.

\begin{table}[tb]
	\caption{Results of the sparse non-local block with different settings on Cityscapes validation set. NoS: number of sampled key elements, GP: global average pooling branch.}
	\label{table:SNL study}
	\centering	
	\begin{tabularx}{\linewidth}{p{4.6cm}  p{0.7cm}<{\centering}  X<{\centering}}
		\hline
		Method               & NoS   & Mean IoU (\%) \\ \hline\hline
		Baseline             & --- & 70.39         \\
		Baseline + SNL         & 9    & 73.54         \\
		Baseline + SNL         & 49   & 74.26         \\
		Baseline + SNL         & 81   & \textbf{74.51}         \\
		Baseline + SNL         & 99  &   73.98       \\
		Baseline + SNL         & 121 & 73.96        \\ \hline
		Baseline + SNL + GP         & 81 & 74.94        \\
		Baseline + SNL + GP + Decoder & 81   & 75.85         \\ \hline
	\end{tabularx}
\end{table}

\begin{table}[t]
	
	\caption{Inference time and accuracy comparisons with other context aggregation approaches on Cityscapes validation set.}
	\label{table:approach compare}
	\centering	
	\begin{tabularx}{\linewidth}{p{2.7cm}  p{2.2cm}<{\centering}  X<{\centering}}
		\hline
		Method                               & Inf. time (ms) & Mean IoU (\%) \\ \hline\hline
		\noalign{\smallskip}
		Baseline & 27             & 70.39         \\
		+ PPM \cite{zhao2017}                 & 41             & 72.88         \\
		+ ASPP \cite{chen2017rethinking}      & 55             & 74.48         \\
		+ NL \cite{wang2018non}               & 42             & 73.75         \\
		+ SNL                                 & 39             & 74.51         \\ \hline
	\end{tabularx}
	
\end{table}

\noindent\textbf{Ablation on global pooling.} As shown in Table~\ref{table:SNL study}, adopting the global average pooling path improved the mean IoU from 74.51\% to 74.94\%, which indicates the effect of this branch.    

\noindent\textbf{Ablation on decoder.} A simple decoder was employed at the end of the network to replace the naive bilinear upsampling operation. Accuracy was improved by around 0.9\% (74.94\% $\rightarrow$ 75.85\%), as shown in Table~\ref{table:SNL study}. This improvement mainly came from the low-level features, which provided spatial information to help refine boundaries.     

\noindent\textbf{Ablation on SNL block.} To compare the performance of our approach with other context aggregation approaches, we replaced the SNL block with the PPM \cite{zhao2017}, the ASPP module \cite{chen2017rethinking} and the standard NL block \cite{wang2018non}, respectively, in the MobileNetV2 backbone, and then we evaluated these models and estimated the inference time with $2 \times 3 \times 1024 \times 2048$ input images on a Titan V100 GPU. The results are reported in Table~\ref{table:approach compare}. It can be seen that our approach outperformed all the other methods with faster inference speed. This  indicates spatial attention is a more effective way to encode multi-scale contextual features, and using part of key elements instead of all can sift out irrelevant contextual information.     
	
\subsection{Comparisons with State-of-the-Art Methods}
Based on the ablation studies, we designed the sparse spatial attention network with ResNet-101 backbone and SNL block, and evaluated it on the Cityscapes, PASCAL Context and ADE20K datasets using multi-scale testing strategy. 

The comparisons with the state-of-the-art methods on the cityscapes test set are shown in Table~\ref{acc}. SSANet outperformed these previous methods with 81.8\% mean IoU. For training on the PASCAL Context and ADE20K datasets, we changed the sampling number in SNL block to 49 to fit the smaller cropping size (i.e., $480\times480$ for PASCAL Context and $520\times520$ for ADE20K). As shown in Table~\ref{table:pascal}, SSANet achieved state-of-the-art performance on both datasets, showing the robustness of our approach.

\begin{table}[t]
	\caption{Segmentation results on Cityscapes test set.} 
	\label{acc}
	\centering	
	\begin{tabularx}{\linewidth}{p{2.7cm}  p{2.2cm}<{\centering}  X<{\centering}}
		\hline
		Method                       & Backbone     & Mean IoU (\%) \\ \hline\hline
		PSPNet \cite{zhao2017}       & ResNet-101   & 78.4          \\
		AAF \cite{ke2018adaptive}    & ResNet-101   & 79.1          \\
		PSANet \cite{zhao2018psanet} & ResNet-101   & 80.1          \\
		DenseASPP \cite{yang2018}    & DenseNet-161 & 80.6          \\
		ANNet \cite{zhu2019asymmetric} & ResNet-101   & 81.3          \\
		CCNet \cite{huang2019ccnet}  & ResNet-101   & 81.4          \\
		DANet \cite{fu2019dual}      & ResNet-101   & 81.5          \\ \hline
		SSANet                       & ResNet-101   & 81.8          \\ \hline
	\end{tabularx}
	
\end{table}

\begin{table}[t]
			\caption{Results on validation sets of PASCAL Context and ADE20K in terms of mean IoU (\%).}
			\label{table:pascal}
			\centering	
			\begin{tabularx}{\linewidth}{p{2.7cm}  p{1.4cm}<{\centering}  X<{\centering}}
				\hline
				Method                         & ADE20K & PASCAL Context \\ \hline\hline
				EncNet \cite{zhang2018context} & 44.65  & 51.7           \\
				DANet  \cite{fu2019dual}       & 45.22  & 52.6           \\
				ANNet \cite{zhu2019asymmetric} & 45.24  & 52.8           \\
				OCRNet \cite{yuan2019object}   & 45.28  & 54.8           \\ \hline
				SSANet                         & 45.50  & 54.9           \\ \hline
			\end{tabularx}
\end{table}
	
\section{Conclusions} 
In this paper, we present the sparse spatial attention network (SSANet) for semantic segmentation. A sparse non-local (SNL) block is proposed and integrated in the network. It utilizes spatial attention mechanism to aggregate multi-scale contextual information and capture long-range dependencies adaptively to improve the performance. Different from the standard non-local block, the dense affinity matrix is replaced with a sparse affinity matrix in the proposed SNL block to improve efficiency and accuracy. The ablation experiments show that SNL block significantly improves segmentation accuracy and achieves the best performance comparing with other context aggregation approaches. SSANet has achieved state-of-the-art results on the Cityscapes, PASCAL Context and ADE20K datasets, demonstrating the benefit and effectiveness of the proposed method.

\bibliographystyle{IEEEbib}
\bibliography{refs}

\end{document}